%% file: acl_revised_2.tex
\pdfoutput=1

\documentclass[11pt]{article}

\usepackage{ACL2023}

\usepackage{times}
\usepackage{latexsym}

\usepackage{booktabs}
\usepackage{graphicx}
\usepackage{amsmath}
\usepackage{amsfonts} 
\usepackage{algorithm} 
\usepackage{algpseudocode} 

\usepackage[T1]{fontenc}

\usepackage[utf8]{inputenc}

\usepackage{microtype}

\usepackage{inconsolata}

\usepackage{multirow}
\usepackage{verbatim}

\usepackage{pifont}
\newcommand{\xmark}{\ding{56}}
\title{Incorporating Graph Information in Transformer-based AMR Parsing}

\author{Pavlo Vasylenko$^{1}$ \qquad Pere-Llu\'is Huguet Cabot$^{1,2}$\thanks{$^*$ Equal contributions.} \\ {\bf
Abelardo Carlos Mart\'inez Lorenzo$^{1,2*}$ \qquad Roberto Navigli$^1$}\\
         $^1$ Sapienza NLP Group, Sapienza University of Rome \\
         $^2$ Babelscape, Rome \\
         \texttt{vasylen.pavlo@gmail.com} \\
         \texttt{\{martinez, huguetcabot\}@babelscape.com} \\
         \texttt{navigli@diag.uniroma1.it}}
         
\begin{document}
\include{utils}
\maketitle
\begin{abstract}

Abstract Meaning Representation (AMR) is a Semantic Parsing formalism that aims at providing a semantic graph abstraction representing a given text. Current approaches are based on autoregressive language models such as BART or T5, fine-tuned through Teacher Forcing to obtain a linearized version of the AMR graph from a sentence. In this paper, we present LeakDistill, a model and method that explores a modification to the Transformer architecture, using structural adapters to explicitly incorporate graph information into the learned representations and improve AMR parsing performance. Our experiments show how, by employing word-to-node alignment to embed graph structural information into the encoder at training time, we can obtain state-of-the-art AMR parsing through self-knowledge distillation, even without the use of additional data. We release the code at {\small \url{http://www.github.com/sapienzanlp/LeakDistill}}. 

\end{abstract}

\section{Introduction}

\begin{figure}[!t]
  \centering
  \includegraphics[width=\columnwidth]{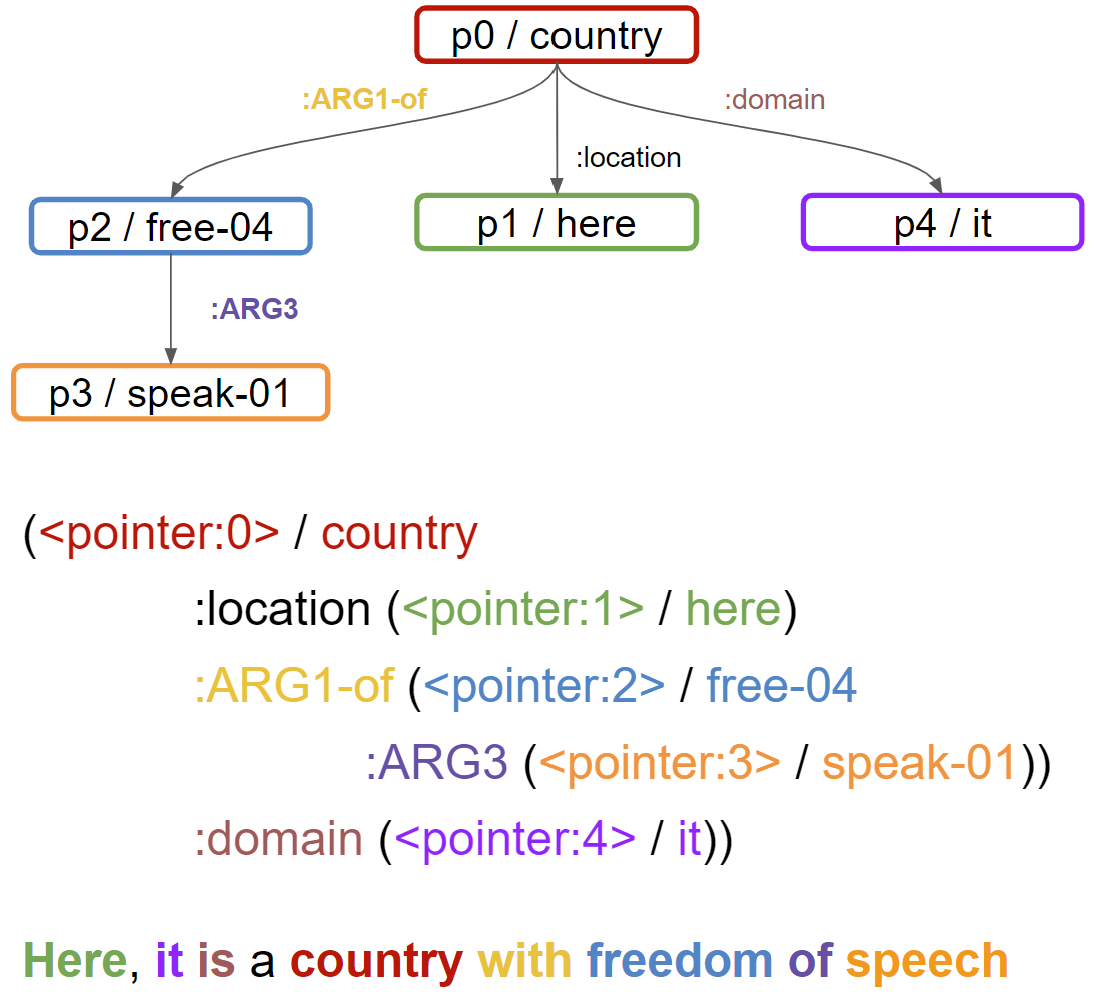}

  \caption{Top: sentence. Middle: AMR graph. Bottom: Linearized graph. Alignment is represented by colours.}
  \label{fig:amr-align-example}
\end{figure}

Creating a machine-interpretable representation of meaning lies at the core of Natural Language Understanding and has been framed as the Semantic Parsing task. Multiple formalisms have been proposed over the years, e.g., Prague Czech-English Dependency Treebank~\cite{hajic-etal-2012-announcing}, Universal Conceptual Cognitive Annotation~\cite{abend-rappoport-2013-universal}, BabelNet Meaning Representation~\citep{bmr-etal-2022-bmr, martinez-lorenzo-etal-2022-fully}; however, Abstract Meaning Representation~\cite[AMR]{banarescu-etal-2013-abstract} has received more attention thanks to the large corpus available and a well-defined structure. AMR captures text semantics in the form of a directed acyclic graph (DAG), with nodes representing concepts and edges representing semantic relationships between them (see Figure \ref{fig:amr-align-example}). Currently, AMR is widely employed in a plethora of NLP domains, such as Information Extraction~\citep{rao-etal-2017-biomedical}, Text Summarization~\citep{hardy-vlachos-2018-guided,liao-etal-2018-abstract}, Question Answering~\citep{lim-etal-2020-know, bonial-etal-2020-infoforager, kapanipathi-etal-2021-leveraging}, Human-Robot Interaction~\citep{bonial-etal-2020-dialogue}, and  Machine Translation~\citep{song-etal-2019-semantic}, among others.

Until a short while ago, autoregressive models proved to be the best approach for semantic parsing because of their outstanding performance without relying on sophisticated ad-hoc architectures~\cite{bevilacqua-etal-2021-one}. Then, more recently, several approaches have emerged to increase performance by including structural information in the model~\cite{naacl-amrize-2021}, adding extra Semantic Role Labeling tasks~\cite{bai-etal-2022-graph} or by ensembling strategies~\cite{neurips-graphene-2021, lee-etal-2022-maximum}.

In this paper, following the effort of strengthening the model's learning phase by incorporating meaningful structural information, we investigate the use of structural adapters~\citep{ribeiro-etal-2021-structural} that are basically Graph Neural Networks (GNNs) embedded in the encoder of a Transformer Encoder-Decoder architecture. The 
 structural information is derived from intrinsic concept-node alignments from which we build a word-based graph with a structure similar to the original AMR. Leveraging such a graph implies partial data leakage: the graph structure is revealed to a model during training. 
 To overcome the lack of the leaked information at inference time, we explore Knowledge Distillation (KD), a technique that transfers knowledge from a teacher model to a student model~\cite{Hinton2015DistillingTK}. The word-based graph is employed with the structural adapters to obtain soft targets (the teacher path), which are then used for self-distillation, transferring the knowledge to the student, which only has access to the text.

Our main contributions are: i) exploring how to add structural information to the AMR parsing model using structural adapters and self-knowledge distillation, ii) state-of-the-art results in AMR parsing for AMR 2.0 and AMR 3.0 datasets, and iii) competitive base models for AMR parsing.

\section{Related Work}

Over the years, multiple trends have appeared to parse AMR graphs: using statistical methods~\citep{flanigan-etal-2014-discriminative, flanigan-etal-2016-cmu, wang-etal-2015-boosting}, neural-transition based parsers~\citep{ballesteros-al-onaizan-2017-amr, liu-etal-2018-amr, fernandez-astudillo-etal-2020-transition, zhou-etal-2021-amr} or bidirectional Transformers~\citep{lyu-titov-2018-amr, zhang-etal-2019-amr, cai-lam-2020-amr} based on BERT~\citep{devlin-etal-2019-bert}.

Recently, autoregressive models based on BART~\cite{lewis-etal-2020-bart} have emerged as a dominant approach for AMR parsing, since they obtained state-of-the-art performance without complex pipelines. One notable example is SPRING~\cite{bevilacqua-etal-2021-one}, which frames AMR parsing as a neural machine translation task, where text is translated into a linearized version of the graph. Subsequently, several works extended SPRING using a variety of different strategies. \citet{procopio-etal-2021-sgl} leverages multitask learning to improve cross-lingual AMR parsing results. ATP~\cite{naacl-amrize-2021} expands the dataset with extra auxiliary tasks such as Semantic Role Labeling and Dependency Parsing, with pseudo-AMR graphs constructed based on a particular task. AMRBART~\cite{bai-etal-2022-graph} uses a pre-training strategy based on Masked Language Modeling where both text and graph need to be denoised, using 200k graphs generated by SPRING. However, despite their efforts to enhance SPRING's performance, all these systems rely on additional external data.  Although Ancestor~\cite{yu-gildea-2022-sequence}, which modifies ancestor information during decoding, and BiBL~\cite{cheng-etal-2022-bibl},  that adds a secondary graph masking task while training, do not rely on extra data, their performance improvements remain relatively limited. Our proposed model effectively bridges the gap in performance between "with" and "without" extra data by integrating explicit structural information during the training phase.

\begin{figure*}[!htp]
  \centering
  \includegraphics[width=\textwidth]{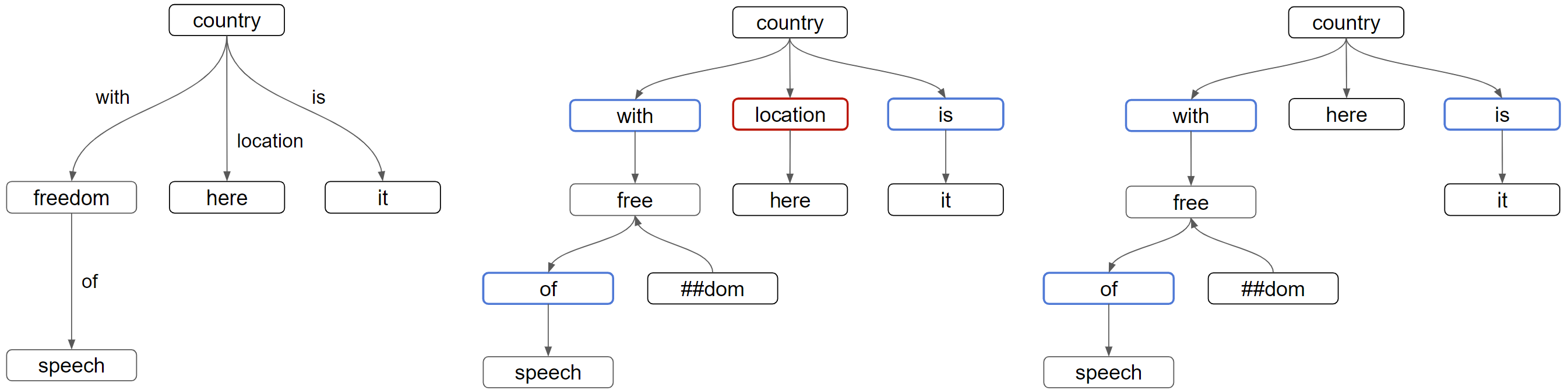}

  \caption{WAG construction of the sentence: "Here, it is a country with the freedom of speech". A graph where AMR concepts are replaced with words (left), a Full WAG (center) and a Contracted WAG (right). Blue lines indicate former AMR relations, and red lines indicate non-aligned nodes. Best seen in color.}
  \label{fig:wag-construction}
\end{figure*}

\section{Word-Aligned Graph}\label{section:wag}

Our goal is to incorporate graph-structured information into the encoder of a Transformer-based parser. However, the model only has access to the input sentence at that stage, with no hidden representation of AMR-specific nodes and relations. Thus, we simplify the AMR structure to a word-based graph by exploiting a pre-existing alignment between spans in text and semantic units in the corresponding AMR graph (see Figure \ref{fig:amr-align-example}). 

First, starting with the source AMR graph, we replace the labels of the AMR nodes and relations with the words of the corresponding sentence as provided by the alignment (Figure \ref{fig:wag-construction}, left).
Next, we convert each edge into a node and connect it to its original endpoints (see Figure \ref{fig:wag-construction}, center).  Moreover, following what \citet{struct-adapt} did for AMR graphs, we split each multi-token node (e.g., \texttt{freedom} in Figure \ref{fig:wag-construction}) into a parent node represented by the first token and children nodes connected to it which contain the remaining tokens. We name the resulting graph representation the Word-Aligned Graph (WAG). 

We will leverage WAGs to enrich the encoder's hidden representations of words with the AMR graph's structural information. Unfortunately, a problem arises with non-aligned nodes (e.g., the \texttt{:location} relation in Figure \ref{fig:wag-construction}), since they will not have associated hidden states. Therefore, we have two alternatives: i) remove nodes for which we do not have hidden states (\textit{Contracted WAG}), or ii) create new hidden states for them (\textit{Full WAG}).

\paragraph{Contracted WAG}\label{wag-contract}
As a first option, we remove non-aligned nodes from the graph. However, deleting the nodes from the original graph would produce a disconnected graph. To obtain a connected structure similar to the original graph, we contract nodes rather than removing them. A contracted WAG (\textit{CWAG}) is a graph in which non-aligned nodes are merged with their closest parent node along with all their relations. Figure \ref{fig:wag-construction} (right) depicts a CWAG. 

\paragraph{Full WAG}\label{align_full_graph}
Alternatively, we preserve the nodes without alignment (e.g., the node ``location'' in Figure \ref{fig:wag-construction} (center)). This type of graph is referred to as a Full WAG (FWAG), Figure \ref{fig:wag-construction} (center) shows an example of FWAG. 
\begin{figure}[!t]
  \centering
  \includegraphics[width=.5\columnwidth]{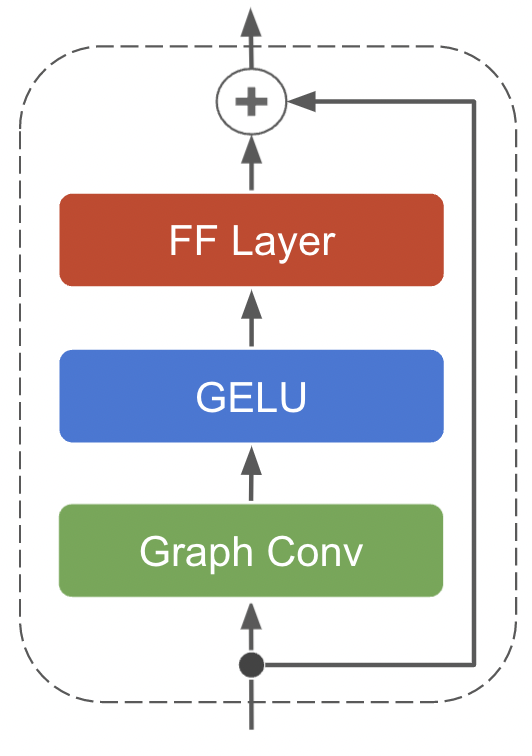}

  \caption{Structural adapter without layer normalization and with GELU activation.}
  \label{fig:struct-adapter}
\end{figure}

\section{Structural Adapters for AMR parsing}
\label{sec:components}

In this  section, we describe the main components of our structure-enhanced approach to AMR parsing. 

\subsection{Parsing with BART}\label{sec:parsing_transformers}
AMR parsing can be defined as a sequence-to-sequence (seq2seq) problem where the input $x=(x_1, ..., x_n)$ is a sequence of $n$ words (or subwords) and the output $g=(e_1, ..., e_m)$ is a linearized graph with $m$ elements. Our goal is to learn a function that models the conditional probability:
\begin{equation}
    p(g|x) = \prod_{t=1}^{m}p(e_t|e_{<t}, x),
\end{equation}
where $e_{<t}$ are the tokens of the linearized graph $g$ before step $t$.

Suppose we have a dataset $D$ of size $|D|$ which consists of pairs $(x^i, g^i)$, with each $g^i$ having length $m^i$. Our objective is then to minimize a negative log-likelihood loss function:
\begin{equation}
\begin{split}
   L_{nll}^D = L_{nll}(D) = -\sum_{i=1}^{|D|}\log p(g^i|x^i) = \\ = -\sum_{i=1}^{|D|}\sum_{t=1}^{m^i}\log p(e_t^i|e_{<t}^i, x^i)
\end{split}
\end{equation}

We use BART as our seq2seq model implementing the above formulation and, following \citet[SPRING]{blloshmi-etal-2021-spring},  add special tokens corresponding to i) AMR-related tokens, ii) variable names <R0>, <R1>, ... <R$n$>, and iii) other tokens needed for the graph linearizations. Then, we fine-tune BART with the input $x$ and the target $g$.

\begin{figure*}[!htp]
  \centering
  \includegraphics[width=\textwidth]{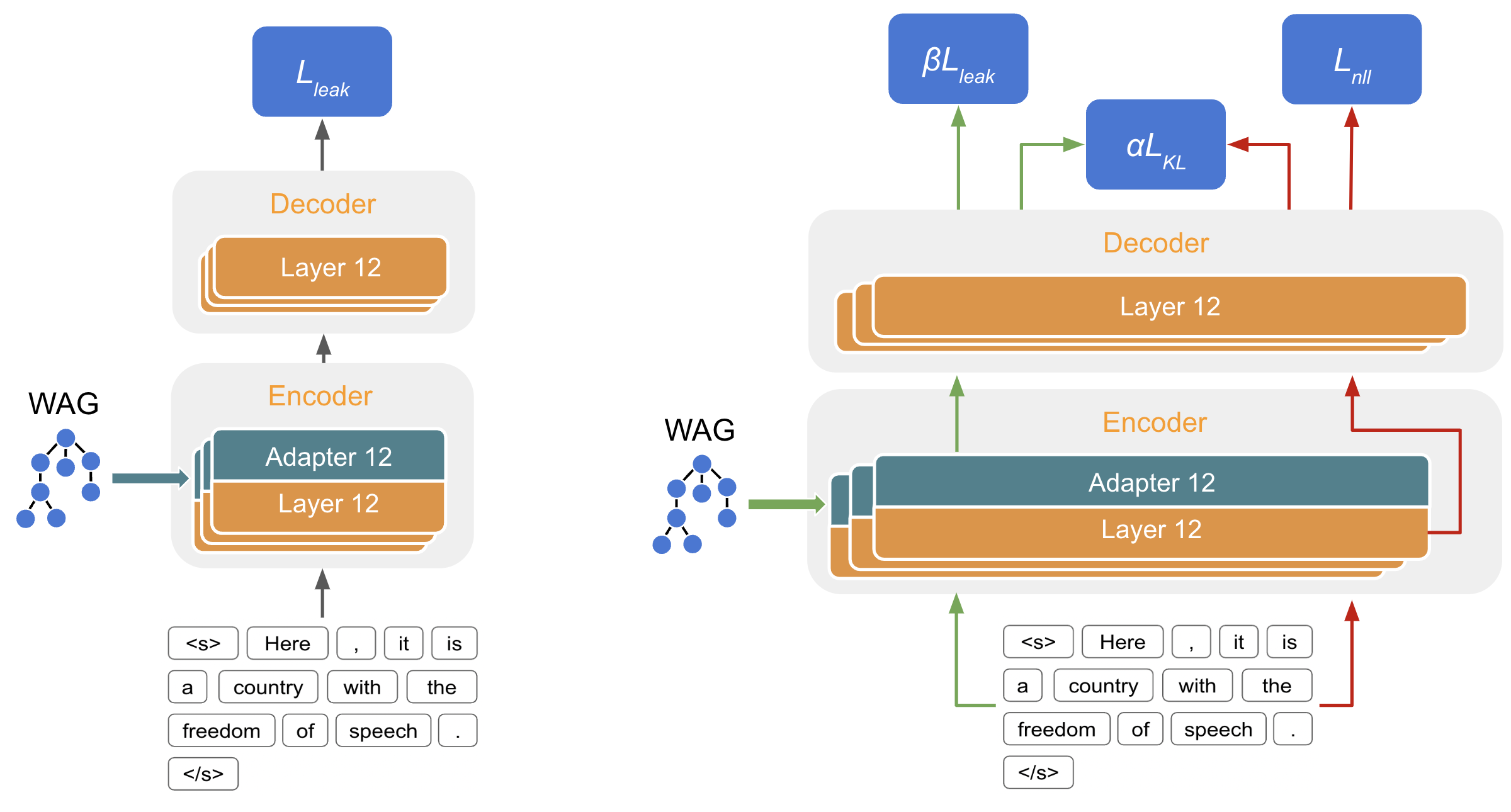}
  \caption{Left: Scheme of the Graph Leakage Model. Right: Scheme of the LeakDistill method with two forward paths:
  the green path incorporates WAG information via adapters; the red path omits adapters, and it is basically the outcome model for the problem. Consequently, the green path is engaged exclusively during the training phase to guide the red path, while during the inference process, only the red path is operative.}
  \label{fig:gnn-encoder}
\end{figure*}

\subsection{Structural adapters}

To incorporate AMR structural information into the encoder, we embed the WAGs -- obtained from AMR graphs as illustrated in Section \ref{section:wag} -- into adapters that encode the graph structure imposed by them. Structural adapters, as introduced by~\citet{struct-adapt}, are a modification of the Transformer architecture that improves pre-trained language models for modeling graph information. They consist of a Graph Convolutional (GraphConv) layer and a feed-forward layer, which are connected through a residual connection. Moreover, we remove layer normalization and set GELU as an activation function (see Figure \ref{fig:struct-adapter}). 

Structural adapters are inserted after each encoder's layer (see Figure \ref{fig:gnn-encoder}). For each hidden representation $\textbf{h}^l_v \in \mathbb{R}^{b}$ from the encoder layer $l$ and the set of edges $\mathcal{E}$ in the WAG, we define the GraphConv operation as:

\begin{equation}
    \text{GraphConv}_l (\textbf{h}^l_v, \mathcal{E}) = \sum_{u\in \mathcal{N}(v)}\frac{1}{\sqrt{d_{u}d_{v}}}\textbf{W}^{l}_g \textbf{h}_{u}^{l}
\end{equation}
where $\mathcal{N}(v)$ is the set of node $v$'s adjacent nodes in the WAG (including $v$ itself), $d_v$ is the degree of $v$, and $\textbf{W}^{l}_g \in \mathbb{R}^{b \times b}$ is a parameter matrix.
Then, the updated hidden states $\textbf{z}_v^l$ are computed as:
\begin{equation}
\begin{split}
\textbf{g}_v^l &= \text{GraphConv}_l (\textbf{h}^l_v, \mathcal{E}) \\ \textbf{z}_v^l &= \textbf{W}^l_a \sigma(\textbf{g}_v^l) + \textbf{h}_v^l,
\end{split}
\end{equation}
where $\sigma$ is the GELU activation function and $\textbf{W}^l_a \in \mathbb{R}^{b \times b}$ is the feed-forward layer parameter matrix.

\section{Our Models}
\subsection{Graph Leakage Model}\label{section:GLM}

We bring together the two main components described in Section \ref{sec:components} by incorporating structural adapters in each layer of the encoder of a BART-based AMR parsing model (see Figure \ref{fig:gnn-encoder} (left) and Algorithm \ref{algo:encoder}). Here, a WAG, together with the hidden representations of tokens in the sentence, are input to the adapters. Since WAGs are constructed using gold AMR graphs, this constitutes a form of information leakage. We name this model the Graph Leakage Model (GLM), with the idea that it will serve as a study of the impact on performance when including WAGs (be they contracted or full, cf. Section \ref{section:wag}). 

To use FWAGs as input to the adapter, we need representations for non-aligned nodes that do not have an associated hidden state. Therefore, for nodes with labels corresponding to AMR special tokens (e.g., \texttt{:location}) we use their embedding. For other nodes, we tokenize the label and take the average embedding. Furthermore, these representations are concatenated after the hidden states in the first adapter layer. After each adapter block, we split representations into two groups: i) the updated hidden states for the original input tokens, which serve as inputs of the subsequent Transformer layer, ii) the updated hidden states for the non-aligned nodes, which are concatenated again in the next adapter block (see Algorithm \ref{algo:encoder}). 

Then, for both CWAG and FWAG, the input to each adapter layer $l$ consists of a matrix of hidden states $H^{l}$ and a set of edges $\mathcal{E}$. 
Note that the set of edges $\mathcal{E}$ does not change through layers. Finally, the loss function for GLM is:
\begin{equation}
    L_{leak} = L_{nll}(\Tilde{D}) = -\sum_{i=1}^{|\Tilde{D}|}\log q(g^i|x^i, w^i),
\end{equation}
where $\Tilde{D}$ is the updated dataset consisting of pairs $((x^i, w^i), g^i)$, $q$ is the probability for GLM, $w^i$ is the WAG. 

\begin{algorithm}[t]
	\caption{Modified BART Encoder}
	\begin{algorithmic}
            \State \textbf{Input:} $\mathcal{E}$ - set of WAG edges, $S^0$ - states for non-aligned nodes, $H^0$ - initial hidden states of the input sequence
		\For {$l  \in \{1, ..., 12\}$ }
            \State $H^l \gets$ BARTLayer$_{l}$($H^{l-1}$)
            \If {Leak Mode}
                \If {Full WAG}
                    \State $G^{l} \gets$ Concat($H^l$, $S^{l-1}$)
                \Else
                    \State $G^{l} \gets$ $H^l$
                \EndIf
                \State $\widetilde{G}^{l} \gets$ StructAdapt$_{l}$($G^{l}$, $\mathcal{E}$)
                \If {Full WAG}
                    \State $[ \widetilde{H}^{l}; S^{l} ] \gets $ Split($\widetilde{G}^{l}$)
                \Else
                    \State $\widetilde{H}^{l} \gets \widetilde{G}^{l}$
                \EndIf
            \Else
                \State $\widetilde{H}^{l} \gets H^l$
            \EndIf
            \State $H^{l} \gets $ $\widetilde{H}^{l}$
		\EndFor
	\end{algorithmic} 
 \label{algo:encoder}
\end{algorithm}

\subsection{Knowledge Distillation}\label{KD_model}

GLM leverages the alignment information to improve the model's understanding of the graph structure and enhance its (the model's) performance in AMR parsing. Unfortunately, as discussed in the previous section, this constitutes a form of leakage at inference time. Therefore, following the idea of Knowledge Distillation \cite[KD]{Hinton2015DistillingTK}, we set the fine-tuned GLM as a teacher model, which receives both the sentence and WAG as inputs, and our plain BART parser as the student (see Section \ref{sec:parsing_transformers}).
Then, the knowledge acquired by the teacher model is  transferred to the student model, which only has access to the sentence. This enables the utilization of WAGs during training while avoiding their use during inference. Hence, our objective is to achieve the following:

\begin{equation}\label{eq:dist_KD}
    p(g|x) = q(g|x, w)
\end{equation}
where $p$ and $q$ are probabilities of the student and the teacher, respectively, and $w$ is the WAG, used only at training time.

As is common in KD, we employ Kullback–Leibler divergence to match the student and the teacher probabilities:

\begin{equation}
    L_{KL} = KL(p, q) = \sum_{k = 0}^{C - 1}{p_k}\log(\frac{p_k}{q_k})
\end{equation}
where $C$ is the number of classes, i.e. our token vocabulary. 
Usually, the loss $L_{nll}^D$ for the original task is added to the total loss, thus becoming:

\begin{gather}
L_{KD} = L_{nll}^D + \alpha L_{KL} = \nonumber\\[1ex]
    = -\sum_{i=1}^{|D|} \sum_{t=1}^{m^k} \sum_{k = 0}^{C - 1}(\delta^i_t(k) \log p^i_{t,k} - \alpha\, {p^i_{t,k}}\log(\frac{p^i_{t,k}}{q^i_{t,k}})), \nonumber\\[1ex]
    p^i_{t,k}=p(e_t^i{=}k \,|\, e_{<t}^i, x^i), \,\,\,\,\,\,\, \nonumber\\
    q^i_{t,k}=q(e_t^i{=}k \,|\, e_{<t}^i, x^i, w^i)
\end{gather}
where $\delta^i_t(k)$ is 1 when $k$ is a target class at step $t$ and 0 otherwise; 
$\alpha$ is a hyperparameter.

There are only architectural differences between the teacher and the student model at the encoder, since the teacher additionally includes the structural adapters. Therefore,  we copy the GLM decoder to the student model and freeze the decoder parameters.

\subsection{LeakDistill}\label{section:2-path-model}
We anticipate that, in our experimentation, KD will have failed to properly transfer the structural information to the student model.
Therefore, we propose a single model approach that can be trained by performing two forward passes at each training step, one with and one without the WAG structural information (see Figure \ref{fig:gnn-encoder} and Algorithm \ref{algo:TPM}). We force the two passes to learn the same distribution by adding a Kullback–Leibler divergence loss to the output logits. As a result, the total loss becomes:

\begin{gather}
L_{LeakDistill} = L_{nll}^D + \beta L_{leak} + \alpha L_{KL} = \nonumber\\[1ex]
= -\sum_{i=1}^{|D|} \sum_{t=1}^{m^k} \sum_{k = 0}^{C - 1}(\delta^i_t(k) \log p^i_{t,k} 
+ \beta\,\delta^i_t(k) \log q^i_{t,k} \nonumber\\
- \alpha\, {p^i_{t,k}}\log(\frac{p^i_{t,k}}{q^i_{t,k}})), \nonumber\\[1ex]
\end{gather}
where $L_{leak}$ is the loss for the first pass (basically, GLM), with leaked information, $L_{nll}^D$ is the loss for the second pass (basically, BART), which is the original negative log-likelihood loss, and finally $L_{KL}$ is the above-described Kullback–Leibler divergence loss. $\alpha$ and $\beta$ are hyperparameters to control each loss scale.

The above formulation implements what is called self-knowledge distillation~\cite[SKD]{hahn-choi-2019-self}. Specifically, in our work we project the knowledge via leveraging data leakage in the first pass rather than computing soft target probabilities. Moreover, we calculate KL divergence for all classes to obtain more knowledge. Finally, based on the intuition that there is not enough information to distill at the beginning of training, we schedule a gradual decrease of $L_{leak}$'s multiplier $\beta$.

\section{Experimental Setup}

To demonstrate the benefits of incorporating structural information in AMR parsing, we devise a set of experiments to assess its performance in comparison to state-of-the-art models. Before delving into details, we provide information regarding the datasets (Section \ref{Datasets}), the metrics (Section \ref{Metric}) and the model (Section \ref{Experiments-Models}) used in our experiments.

\subsection{Datasets}\label{Datasets}

We test on two AMR benchmark datasets: i) AMR 2.0, which has 36521, 1368, and 1371 sentence-AMR pairs in the training, validation, and test sets, respectively, and ii) AMR 3.0, which contains 55635, 1722, and 1898 sentence-AMR pairs in the training, validation, and test sets, respectively (see Appendix \ref{appendix:data}). Furthermore, we test on The Little Prince (TLP) and the Bio AMR out-of-distribution datasets.

\paragraph{Alignment}

Our approach relies directly on the structural information extracted from the word-concept alignment. There are several alignment standards: first, Information Sciences Institute (ISI) provides extended AMR 2.0 and AMR 3.0 datasets with alignments of all the graph semantic units that are directly related to the sentences' spans~\cite{pourdamghani-etal-2014-aligning}. Second, Linguistically Enriched AMR~\citep[LEAMR]{blodgett-schneider-2021-probabilistic} achieves full graph-alignment coverage by aligning all the graph semantic units to a corresponding span in the sentence.

\paragraph{Silver Data} 
Following~\citet{bevilacqua-etal-2021-one}, we explore the same strategy to generate a dataset with 140k silver sentence-graph pairs. The silver LEAMR alignments are generated using the approach of \citet{amr-alignment-2022}.
\subsection{Metrics}\label{Metric}  We evaluate our models using the SMATCH metric (see Appendix \ref{appendix:metric} for more details). Additionally we also perform evaluation with two additional metrics: S$^{2}$MATCH \cite{opitz-etal-2020-amr} and WWLK \cite{opitz-etal-2021-weisfeiler}. For WWLK we use WWLK-k3e2n introduced in \citet{opitz-etal-2021-weisfeiler}.
\subsection{Models}\label{Experiments-Models}
We use SPRING~\citep{bevilacqua-etal-2021-one} as our baseline, and an auto-regressive model based on BART~\citep{lewis-etal-2020-bart} for predicting linearized versions of AMR graphs. Our models are built on top of this model, inheriting
some hyperparameters (see Table \ref{table:hp-space}). 

In order to address the issue of overfitting, we implement a masking strategy which is used in conjunction with dropout and weight decay. For each batch, input tokens are masked with a varying probability $p_{mask}$, which is uniformly sampled from the specified masking range (see Appendix \ref{sec:appendix} for details). The strategy is used for all models including SPRING (ours). In the following paragraphs, we explain the specific setup per each model.

\begin{table}[!t]
\centering
\begin{tabular}{lc}
\toprule
\textbf{Model} &  \textbf{AMR 3.0} \\
\midrule
SPRING (ours) & 84.55 \\
\midrule
Contracted WAG & 86.01 \\
Full WAG & 89.58  \\
\bottomrule
\end{tabular}
\caption{GLM results for AMR 3.0 development set. }
\label{table:GLM}
\end{table}

\begin{table}[t!]
\centering
\begin{tabular}{lcc}
\toprule
\textbf{} & \textbf{Model} &  \textbf{AMR 3.0} \\
\midrule
& SPRING (ours) & 84.55 \\
\midrule
\multirow{1}{*}{KD} 
                       & Full WAG (89.58) & 83.90 \\
\midrule
\multirow{3}{*}{\shortstack[l]{LeakDistill \\ (Self-KD)}} 
& $L_{leak}$ + $L_{nll}^D$ & 84.47 \\
& $L_{leak}$ + $L_{KL}$  & 85.03 \\
& $L_{leak}$ + $L_{nll}^D$ + $L_{KL}$  & \textbf{85.04} \\
 \bottomrule
\end{tabular}
\caption{Knowledge Distillation results for the development set of AMR 3.0.}
\label{table:KD}
\end{table}

\begin{table*}[hbt!]
\resizebox{\textwidth}{!}{
\begin{tabular}{ccc|cccccccc}
\toprule
   \textbf{Model} & \multicolumn{1}{l}{\textbf{Extra Data}} & \textbf{Smatch} & \textbf{Unlab.} & \textbf{NoWSD} & \textbf{Conc.} & \textbf{Wiki} & \textbf{NER}  & \textbf{Reent.} & \textbf{Neg.} & \textbf{SRL}  \\  \midrule \midrule
 SPRING (ours)            & \xmark                                     & 84.4           & 87.4            & 84.8           & 90.4           & \underline{84.1}          & 90.9 & 71.6   & 73.5          & 80.1 \\
 BiBL            & \xmark                                    & 84.6   & 87.8   & 85.1  & 90.3  & 83.6 & 92.5          & 74.4            & 73.9 & 83.1         \\ 
 Ancestor            & \xmark                                    & 84.8   & 88.1   & 85.3  & 90.5  & \underline{84.1}  & 91.8          & \underline{75.1}             & 74.0 & 83.4          \\
 \textbf{LeakDistill}            & \xmark                                    & ~~~~85.7$^{s,o}$   & 88.6   & 86.2  & 91.0  & 83.9 & 91.1          & 74.2            & \textbf{76.8} & 81.8          \\  \midrule
 SPRING         & 200K                                    & 84.3            & 86.7            & 84.8           & 90.8           & 83.1          & 90.5          & 72.4            & 73.6          & 80.5          \\
 ATP            & ~~40K                                     & ~~85.2$^{s}$           & 88.3            & 85.6           & 90.7           & 83.3          & \textbf{93.1} & 74.7   & 74.9          & \textbf{83.3} \\
 AMRBART        & 200K                                    & ~~85.4$^{s}$            & 88.3            & 85.8           & 91.2           & 81.4          & 91.5          & 73.5            & 74.0          & 81.5          \\
\textbf{LeakDistill}          & 140K                                    & ~~~~~~~~~\textbf{86.1}$^{s,o,b,a}$   & \textbf{88.8}   & \textbf{86.5}  & \textbf{91.4}  & 83.9 & 91.6         & \underline{75.1}            & 76.6 & 82.4          \\ \bottomrule
\end{tabular}}
\caption{AMR 2.0 results and comparisons with previous systems. Bold indicates best performance per set, underline in case of a tie. Breakdown extra scores after vertical line. Superscript indicates the result is significantly better using an approximate randomization test~\cite{riezler-maxwell-2005-pitfalls} at $p<0.05$ with respect to $s = SPRING$, $o = SPRING (ours)$, $b = BiBL$, $a = ATP$. We are unable to test Ancestor due to no public checkpoint. Appendix \ref{appendix:metric} contains the descriptions for the columns.}
\label{table:overall-test-amr2}
\end{table*}

\begin{table*}[hbt!]
\resizebox{\textwidth}{!}{
\begin{tabular}{ccc|cccccccc}
\toprule
   \textbf{Model} & \multicolumn{1}{l}{\textbf{Extra Data}} & \textbf{Smatch} & \textbf{Unlab.} & \textbf{NoWSD} & \textbf{Conc.} & \textbf{Wiki} & \textbf{NER}  & \textbf{Reent.} & \textbf{Neg.} & \textbf{SRL}  \\  \midrule \midrule
 SPRING         & \xmark                                    & 83.0           & 85.4            & 83.5           & 89.5           & 81.2         & 87.1          & 71.3            & 71.7          & 79.1          \\
 SPRING (ours)            & \xmark                                     & 83.8           & 86.7            & 84.3           & 89.9           & 81.5          & 87.2 & 71.4   & 71.5          & 79.8 \\
Ancestor           & \xmark                                    & 83.5      & 86.6      & 84.0     & 89.5  & 81.5          &  88.9          & \textbf{74.2}            & 72.6 & 82.2          \\   
BiBL           & \xmark                                    & ~~83.9$^{s}$      & 87.2      & 84.3     & 89.8  & \textbf{83.7}          & \textbf{93.2}          & 73.8           & 68.1 & 81.9          \\   
\textbf{LeakDistill}           & \xmark                                     & ~~~~~~~84.5$^{s,o,a}$      & \underline{87.5}      & \underline{84.9}     & 90.5  & 80.7          &  88.5         & 73.1            & 73.7 & 80.7          \\  \midrule
 ATP            & ~~40K                                     & ~~83.9$^{s}$           & 87.0            & 84.3           & 89.7           & 81.0 & 88.4          & 73.9   & \textbf{73.9}          & \textbf{82.5} \\
AMRBART        & 200K                                    & ~~~~~~~84.2$^{s,o,a}$      & 87.1      & 84.6     & 90.2           & 78.9          & 88.5 & 72.4            & 72.1          & 80.3          \\
\textbf{LeakDistill}           & 140K                                    & ~~~~~~~~~\textbf{84.6}$^{s,o,b,a}$      & \underline{87.5}      & \underline{84.9}     & \textbf{90.7}  & 81.3          & 87.8          & 73.4            & 73.0 & 80.9         \\ \bottomrule
\end{tabular}}
\caption{AMR 3.0 results and comparisons with previous systems. Bold indicates best performance per set, underline in case of a tie. Breakdown extra scores after vertical line. Superscript indicates the result is significantly better using an approximate randomization test~\cite{riezler-maxwell-2005-pitfalls} at $p<0.05$ with respect to $s = SPRING$, $o = SPRING (ours)$, $b = BiBL$, $a = ATP$. We are unable to test Ancestor due to no public checkpoint. Appendix \ref{appendix:metric} contains the descriptions for the columns.}
\label{table:overall-test-amr3}
\end{table*}

\begin{table}[t]
\centering
\begin{tabular}{lcc}
\toprule
\textbf{Model} & \textbf{AMR 3.0} \\
\midrule
SPRING (ours) & 84.55 \\
\midrule
Contracted WAG & 84.90  \\
Full WAG & 85.04  \\
\quad + $\beta$ scheduling &  85.08 \\
\quad+ Silver  & \textbf{85.34} \\
\quad+ Silver + $\beta$ scheduling & 85.28 \\
\midrule
The green path (Figure \ref{fig:gnn-encoder}) \\ FWAG + Silver  & 86.09 \\
\bottomrule
\end{tabular}
\caption{Performance of LeakDistill models on the development set of AMR 3.0. }
\label{table:paerfomance-2-path-model}
\end{table}

\paragraph{Graph Leakage Model}
We explore two different settings for GLM: i) Contracted WAG, and ii) Full WAG (see Section \ref{section:wag}).

\paragraph{Knowledge Distillation} 
We test KD on the GLM with the highest SMATCH among CWAG and FWAG (see Table \ref{table:GLM}).

\paragraph{LeakDistill}
As done for GLM, we first examine the difference in performance between Contracted WAG and Full WAG. Then, we test Full WAG with i) $\beta$ scheduling, ii) the silver data, iii) the combination of the silver data and the $\beta$ scheduling. In the case of the scheduling of $\beta$, we start from $\beta=90$ and decrease it linearly at each iteration for 21k iterations in total until it reaches 10. The hyperparameter $\alpha$ is set to 20. The value of $\beta$ for the case i) and other hyperparameters are listed in Table \ref{table:hp-space}.

\section{Results}\label{section:results}
In this section, we provide our experimental findings. All tables show single-run results.
\paragraph{Graph Leakage Model} Table \ref{table:GLM} shows results for the Graph Leakage Model. While this setup relies on information being leaked from the final graph structure, it sets an upper bound on how encoding such information can improve performance. Here, we observe an increase of around five SMATCH points when using FWAG, whereas CWAG improvements are much smaller. While the model is certainly taking advantage of the leaked information, this is provided through the hidden states of the encoder. Therefore, we need to explore whether some of this performance gain can be kept implicitly without any information leak. Moreover, it is necessary to investigate the persistence of any performance disparity between CWAG and FWAG. This information is intriguing, as CWAG and FWAG differ in the context of additional information availability. CWAG only possesses a structure akin to the original graph, while FWAG not only exhibits a greater degree of structural similarity but also includes the original labels for non-aligned nodes.

\paragraph{KD and LeakDistill} Table \ref{table:KD} compares the results between applying KD with GLM as the teacher versus the LeakDistill approach, explained in Section \ref{section:2-path-model}.We see how KD alone falls short of taking full advantage of the performance gains of GLM. On the other hand, LeakDistill, especially when including the KL loss, leads to about a 0.5 SMATCH point increase on the development set. Hence, we focus on LeakDistill as our main approach. Table \ref{table:paerfomance-2-path-model} shows a breakdown of the experiments with LeakDistill, such as scheduling the KL loss or adding a silver data pretraining phase. It is evident that the performance difference between CWAG and FWAG remains, paving the way for more in-depth research into the types of information that prove advantageous for LeakDistill.
Additionally, the final row of Table \ref{table:paerfomance-2-path-model} presents the outcome when the adaptors are active (the green path). It is noticeable that, despite the green path essentially being the GLM, it fails to match the performance level of 89.58.

\paragraph{Main results} Tables  \ref{table:overall-test-amr2} and \ref{table:overall-test-amr3} shows results for our proposed model, based on BART-large. Our system performs better than any previous single model parser, and, most notably, does so even without extra data, i.e. silver sentence-graph pairs. For AMR 2.0, we see up to 0.7 SMATCH increase over AMRBART and 0.4 on AMR 3.0. The use of extra data only leads to a small improvement, showing the efficiency of our approach, which is able to outperform previous state-of-the-art systems that relied on up to 200K extra samples. In the breakdown performance, we see how our system performs worse than ATP on Reentrancies, Negation and notably SRL. We believe this is due to the multitask nature of ATP, where SRL is explicitly included as a task. This opens the door to future work exploring the interaction between our approach and the inclusion of auxiliary tasks.

It is worth noting that our system relies on alignment information which is openly discussed at various stages in the paper. We do not consider this information as extra data since it is generated based on the existing data.

\paragraph{Out-of-distribution evaluation} Table \ref{table:OOD} shows the Out-of-Distribution of LeakDistill. We see a smaller improvement on TLP, 0.3 over AMRBART. On the harder BioAMR, performance increased by over a point, showing how the model is able to generalize well on different domains.

\paragraph{BART base} Our state-of-the-art system relies on BART-large, which has 400M parameters. While it shows very strong performance, it has a big computational footprint, especially at inference time due to its auto-regressive generative nature. This makes the need for lighter, more compute efficient models an important step towards better Semantic Parsers. Table \ref{table:bart-base} shows the performance of our approach when trained on top of BART-base, which has 140M parameters, achieving 83.5 SMATCH points on AMR 3.0, 1 point higher than AMRBART and,  noticeably, surpassing SPRING-large performance by half a point. We believe it is crucial to have close to state-of-the-art performance base models, closing the gap from 2 points to 1 when compared to their large counterparts.

\paragraph{Other metrics} Recent studies have shown that achieving a higher SMATCH score does not necessarily result in better performance of an AMR parser, as demonstrated by \citet{opitz-frank-2022-better}. To address this issue, we use two additional evaluation metrics, namely S$^2$MATCH and WWLK-k3e2n (WWLK), which measure graded concept similarity and edge label importance, respectively. Our experiments reveal that S$^2$MATCH correlates well with SMATCH, as expected for monolingual parsers. Conversely, WWLK is specifically designed for monolingual AMR parsing and emphasizes edge labels. Interestingly, our findings suggest that ATP performs well, second only to our proposed system, LeakDistill. This may be due to the fact that both systems place greater emphasis on edges, with ATP leveraging semantic role labeling data and LeakDistill utilizing structural information such as edges in the FWAGs. In contrast, AMRBART and BiBL exhibit a significant drop in performance compared to the SPRING baseline, possibly due to their use of masking as an additional signal, as their masking strategies may not be beneficial for edge labels.

\begin{table}[t!]

\centering
\begin{tabular}{lcc}
\toprule
\textbf{Model} &  \textbf{TLP} &  \textbf{BioAMR} \\
\midrule
SPRING & 81.3 & 61.6 \\
BiBL & 78.6 & 61.1 \\
ATP & 78.9 & 61.2 \\
AMRBART & 82.3 & 63.4 \\
\textbf{LeakDistill} & \textbf{82.6} & \textbf{64.5} \\
 \bottomrule
\end{tabular}
\caption{Out of distribution results. AMRBART and SPRING are taken from~\citet{lee-etal-2022-maximum}.}
\label{table:OOD}
\end{table}

\begin{table}[t!]

\centering
\begin{tabular}{lcc}
\toprule
\textbf{Model} &  \textbf{AMR 2.0} &  \textbf{AMR 3.0} \\
\midrule
SPRING & 82.8 & - \\
AMRBART & 83.6 & 82.5 \\
\textbf{LeakDistill} & \textbf{84.7} & \textbf{83.5} \\
\bottomrule
\end{tabular}
\caption{BART-base versions performance.}
\label{table:bart-base}
\end{table}

\begin{table}[t!]

\resizebox{\columnwidth}{!}{%
\centering
\begin{tabular}{lcccc}
\toprule
\textbf{Model} &  \textbf{SMATCH} &  \textbf{\textsc{S$^{2}$MATCH}} & \textbf{\textsc{WWLK}} \\
\midrule
SPRING & 83.0 & 84.2 & 84.8 \\
BiBL & 83.9 & 84.6 & 82.3 \\
ATP & 83.9 & 84.7 & 85.7 \\
AMRBART & 84.2 & 85.1 & 83.9 \\
\textbf{LeakDistill} & \textbf{84.6} & \textbf{85.5} & \textbf{85.9} \\
\bottomrule
\end{tabular}}
\caption{Performance on AMR 3.0 for different metrics. S$^2$MATCH is taken from \citet{opitz-etal-2020-amr}. We use WWLK-k3e2n as proposed in \citet{opitz-frank-2022-better}.}
\label{table:more-metrics}

\end{table}

\section{Performance Analysis}

\begin{figure}[t!]
    \centering
    \includegraphics[width=0.95\columnwidth, trim={1cm 0.5cm 0.5cm 0.5cm},clip]{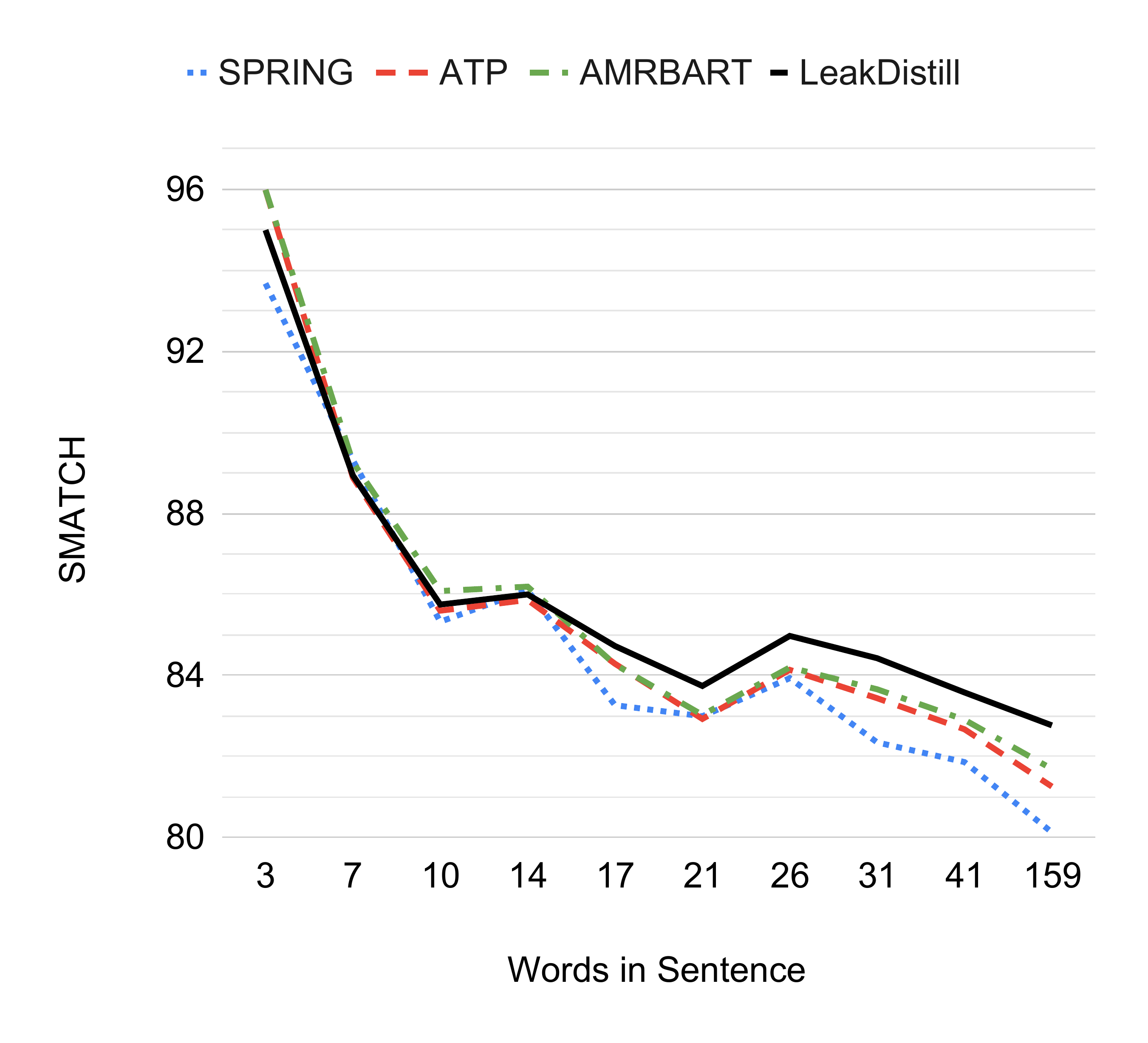}
    \caption{SMATCH score for buckets of 200 instances. X axis shows max. number of words per sentence.}
    \label{fig:error_sent_length}
\end{figure}

Seq2seq parsers show decreased performance for longer sentences since a single error at decoding time in an early step can lead to compound errors and suffer from exposure bias. We explore how this affects our model compared to SPRING, ATP and AMRBART. Figure \ref{fig:error_sent_length} shows the performance on AMR 3.0 test set for buckets of 200 sentences split by the number of words. While performance is similar on shorter sentences, with AMRBART showing slightly better performance, in longer sentences of over 14 words LeakDistill fares better, especially compared to the baseline, which drops to 80 SMATCH points. This experiment also shows how performance is relatively stable for medium-length sentences (10-30 words, oscillating around 85 points), while it starts deteriorating for longer ones. The high performance on short sentences is likely due to easy-to-parse structures, such as single date sentences.

\section{Conclusion}
We presented a new approach to training the Transformer architecture where partial information of the target sequence can be learned via self-knowledge distillation: the information can be leaked in the encoder implicitly through Transformer adapters which improve training but are switched off during inference.
By employing this approach in AMR parsing, we achieved state-of-the-art results among non-ensemble methods. Moreover, we produced a lightweight AMR parser that outperforms SPRING while having four times fewer parameters.
We also showed that, for all methods, performance degrades as the number of words increases.

Interestingly, our approach can potentially be used in other tasks, such as Relation Extraction, where alignments between input and target sequence elements exist, or structural information is unavailable at inference time.

\section{Limitations}
Our approach for training the Transformer architecture using self-knowledge distillation is promising, but there are still some limitations that need to be addressed in future work. One limitation is that our approach is only tested on the task of AMR parsing, and more evaluations are needed to see if it generalizes well to other tasks, such as Relation Extraction. Additionally, our approach, as is also the case for other current methods, exhibits performance degradation as the number of words in the sentence increases. This may be an indication of the current methods' limitation or lack of robustness to longer sentences.

Another limitation is the added complexity and extra parameters required by the use of Transformer adapters, which increases the overall complexity of the architecture and training time. Even though our approach still achieves state-of-the-art results and it is as lightweight as previous systems at inference time, this fact should be considered by researchers if they should decide to adopt it for other tasks.

In summary, our approach presents an innovative way to train the Transformer architecture and achieve state-of-the-art results in AMR parsing. However, more work is needed to further improve the performance of the model and to apply it to other tasks as well.

\section{Ethical considerations}

In considering the ethical and social implications of our proposed approach to AMR parsing, we acknowledge that there are several important considerations to take into account.

One significant concern is the potential for bias in the training data and models, which can result in unfair or discriminatory outcomes for certain groups of individuals. Additionally, the training and test data may not be representative of the population that the model will be applied to, potentially leading to poor performance in specific domains.

Furthermore, our approach relies on the use of Transformer-based models, which have been shown to perpetuate societal biases present in the data used for training. It is, therefore, crucial to ensure that the data used for training is diverse and unbiased.

Moreover, the use of techniques such as self-knowledge distillation may lead to data leakage, where the model overfits the training data and performs poorly on new data, which could have negative impacts on the predictions.

In conclusion, even if we consider our approach does not have negative implications, it is important to note that bias and fairness are complex issues that require ongoing attention and improvement.

\section*{Acknowledgments}

\begin{center}
\noindent
\begin{minipage}{0.1\linewidth}
    \raisebox{-0.25\height}{\includegraphics[trim =0mm 5mm 5mm 5mm,clip,scale=0.045]{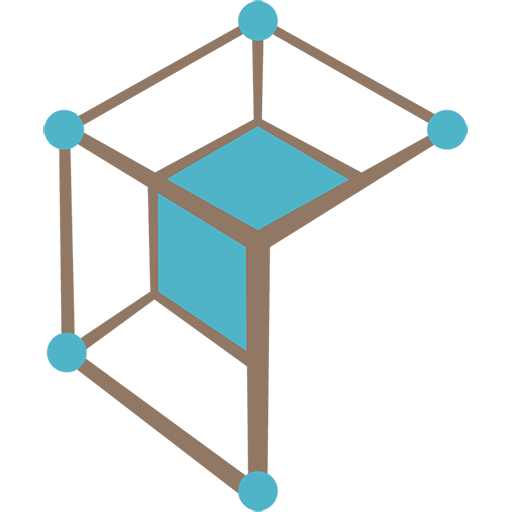}}

\end{minipage}
\hspace{0.005\linewidth}
\begin{minipage}{0.72\linewidth}
The authors gratefully acknowledge the support of the European Union’s Horizon 2020 research project \textit{Knowledge Graphs at Scale} (KnowGraphs) under the Marie  Marie Sk\l{}odowska-Curie grant agreement No \href{https://cordis.europa.eu/project/id/860801}{860801}.

  \vspace{1ex}
\end{minipage}
\hspace{0.005\linewidth}
\begin{minipage}{0.1\linewidth}
  \vspace{0.05cm}
\raisebox{-0.25\height}{\includegraphics[trim =0mm 5mm 5mm 5mm,clip,scale=0.060]{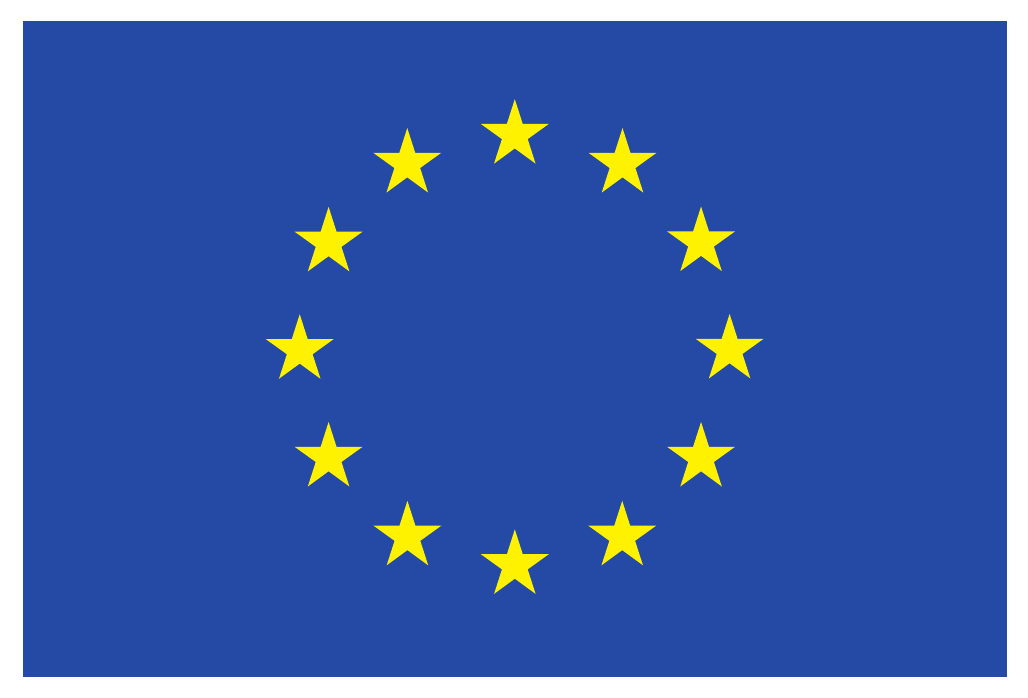}}
  \vspace{0.05cm}
\end{minipage}\\
\end{center}

The last author gratefully acknowledges the support of the PNRR MUR project PE0000013-FAIR.

\bibliography{anthology,custom}
\bibliographystyle{acl_natbib}

\appendix

\newpage

\newpage
\section*{Appendices}

\section{Model Hyperparameters}
Table \ref{table:hp-space} lists hyperparameters and search space for the experiments:
\begin{itemize}
\item LR sched. - learning rate scheduler
\item KL temp. - Kullback–Leibler divergence temperature 
\item AMR 3 aligns. - type of alignments for AMR 3.0
\item Mask. range - masking range. For each batch, we mask the input tokens with probability $p_{mask}$, the value for which is sampled uniformly from the masking range. For instance, the [0; 0.15] range means $p_{mask} \sim \emph{U}(0, 0.15) $
\end{itemize}

The LeakDistill experiments detailed in Table \ref{table:KD} were performed utilizing the final set of hyperparameters listed in Table \ref{table:hp-space}. However, it should be noted that the experiment that did not involve KL loss did not necessitate the use of the variable $\alpha$.

\label{sec:appendix}
\begin{table}[hbt!]
\centering
\resizebox{\columnwidth}{!}{
\begin{tabular}{lcc}
\toprule
\textbf{Group} & \textbf{Parameter} &  \textbf{Values} \\
\midrule
\multirow{7}{*}{\shortstack[l]{Inherited \\ (SPRING)}}
                       & Optimizer & RAdam \\
                       & Batch size & 500 \\
                       & Dropout & 0.25 \\
                       & Attent. dropout & 0 \\
                       & Grad. accum. & 10 \\
                       & Weight decay & 0.004 \\    
                       & LR & 0.00005 \\
                       & Beamsize & 5 \\
\midrule
\multirow{3}{*}{SPRING (ours)} 
                      & LR sched. & const., \textbf{linear} \\
                      & Mask. range  & [0; \{0, \textbf{0.15}\}] \\
                      & Beamsize & 5, \textbf{10} \\
\midrule
\multirow{3}{*}{Adapter} 
                       & Encoder layers & 1-12 \\
                       & Activation & GELU \\
                       & Dropout & \textbf{0.01}, 0.1 \\
\midrule
\multirow{3}{*}{GLM} 
                       & LR & 0.00005, \textbf{0.0001} \\
                       & LR sched. & const., \textbf{linear} \\
                       & Mask. range  & [0; 0.15] \\
\midrule
\multirow{6}{*}{KD} 
                       & $\alpha$ & 10 \\
                       & LR & 0.00005, \textbf{0.0001} \\
                       & LR sched. & const., \textbf{linear} \\
                       & Weight decay & 0.004, \textbf{0.0001} \\
                       & Decoder & train, \textbf{freeze} \\
                       & Mask. range  & [0; 0.15] \\
\midrule
\multirow{5}{*}{LeakDistill} 
                       & LR sched. & const., \textbf{linear} \\
                       & KL temp. & \textbf{1}, 2 \\
                       & $\alpha$ & 1, 5, 10, \textbf{20} \\
                       & $\beta$ & 1, 5, 10, \textbf{sched.} \\
                       & AMR 3 aligns.  & ISI, \textbf{LeAMR} \\
                       & Mask. range  & [0; \{0, 0.1, \textbf{0.15}\}] \\
                       & Beamsize & 5, \textbf{10}\\
 \bottomrule
\end{tabular}}
\caption{Final hyperparameters and search space for the experiments. All groups have the same parameters as original SPRING if they are not overwritten. For instance, SPRING (ours) and for LeakDistill have the same learning rate of 0.00005.  }
\label{table:hp-space}
\end{table}

\section{Hardware and size of the model}
We performed experiments on a single NVIDIA 3090 GPU with 64GB of RAM and Intel®
Core™ i9-10900KF CPU.
The total number of trainable parameters of LeakDistill is 434,883,596.
Training the model on the silver data took 33 hours, whereas further fine-tuning took 16 hours.

\section{BLINK}
All systems from Tables \ref{table:overall-test-amr2} and  \ref{table:overall-test-amr3} use BLINK~\cite{wu-etal-2020-scalable} for wikification. For this purpose, we used the $blinkify.py$ script from the SPRING repository.

\section{Metric}\label{appendix:metric}

We evaluate AMR parsing using the SMATCH metric~\cite{cai-knight-2013-smatch} and extra scores of~\citet{damonte-etal-2017-incremental}: i) Unlabel, compute on the predicted graphs after removing all edge labels, ii) No WSD, compute while ignoring Propbank senses (e.g., duck-01 vs duck-02), iii) Wikification, F-score on the wikification (:wiki roles), iv) NER, F-score on the named entity recognition (:name roles), v) Negations, F-score on the negation detection (:polarity roles), vi) Concepts, F-score on the concept identification task, vii) Reentrancy, computed on reentrant edges only, viii) Semantic Role Labeling (SRL), computed on :ARG-i roles only.

\section{Data}\label{appendix:data}
The AMR 3.0 \cite{amr-annotation-3.0} data used in this paper is licensed under the \textit{LDC User Agreement for Non-Members} for LDC subscribers, which can be found \href{https://catalog.ldc.upenn.edu/LDC2020T02}{here}. The \textit{The Little Prince} Corpus can be found \href{https://amr.isi.edu/download.html}{here} from the Information Science Institute of the University of Southern California.

\section{Algorithms}\label{apendix:algorithms}

Algorithm \ref{algo:TPM} shows one training step of the LeakDistill model.

\begin{algorithm}
\caption{One training step of the LeakDistill model}
    \begin{algorithmic}
     \State \textbf{Input:} X - batch of input sequences and WAGs, Y - batch of target graphs
     
      \State Set Model to Normal Mode
     \State $L_{nll}^D, Probs_{1} \gets Model(X, Y)$ 
     \State Set Model to Leak Mode
     \State $L_{leak}, Probs_{2} \gets Model(X, Y)$
     \State $L_{KL} \gets$ KLDiv $(Probs_{1}, Probs_{2})$
     \State $L \gets \alpha L_{KL} + \beta L_{leak} + L_{nll}^D$
     \State Optimization step of $L$ 
\end{algorithmic} 
 \label{algo:TPM}
\end{algorithm} 

\end{document}

%% file: utils.tex
\newcommand{\hole}[1]{\textcolor{red}{TODO: #1}}
\newcommand{\reviewed}[1]{\textcolor{purple}{#1}}